\documentclass[letterpaper, 10 pt, conference]{ieeeconf}  

\IEEEoverridecommandlockouts                              

\usepackage{graphics} 
\usepackage{epsfig} 
\usepackage{mathptmx} 
\usepackage{times} 
\usepackage{amsmath} 
\usepackage{amssymb}  
\usepackage[utf8]{inputenc}
\usepackage{array}
\usepackage{multirow}

\title{\LARGE \bf
Development of a Compact Robust Passive Transformable
 Omni-Ball for Enhanced Step-Climbing and Vibration Reduction
}

\author{Kazuo Hongo$^{1}$, Takashi Kito$^{1}$, Yasuhisa Kamikawa$^{1}$, Masaya Kinoshita$^{1}$ and Yasunori Kawanami$^{1}$
\thanks{$^{1}$Sony Group Corporation,
 Minato-ku, Tokyo, Japan, 108-0075
{\tt\small email:kazuo.hongo@sony.com}}%
}

\begin{document}

\maketitle
\thispagestyle{empty}
\pagestyle{empty}

\begin{abstract}

        This paper introduces the Passive Transformable Omni-Ball (PTOB), an advanced omnidirectional wheel engineered to enhance step-climbing performance, incorporate built-in actuators, diminish vibrations, and fortify structural integrity. By modifying the omni-ball's structure from two to three segments, we have achieved improved in-wheel actuation and a reduction in vibrational feedback. Additionally, we have implemented a sliding mechanism in the follower wheels to boost the wheel's step-climbing abilities. A prototype with a 127 mm diameter PTOB was constructed, which confirmed its functionality for omnidirectional movement and internal actuation. Compared to a traditional omni-wheel, the PTOB demonstrated a comparable level of vibration while offering superior capabilities. Extensive testing in varied settings showed that the PTOB can adeptly handle step obstacles up to 45 mm, equivalent to 35 $\%$ of the wheel's diameter, in both the forward and lateral directions. The PTOB showcased robust construction and proved to be versatile in navigating through environments with diverse obstacles.

\end{abstract}

\section{Introduction}

Omnidirectional mobility mechanisms have been developed in various types due to their ability to achieve efficient and quick movement without the need for the body to change direction, as well as their ease of movement in narrow spaces. One of the simplest types of omnidirectional wheels is equipped with actuators that allow steering in the yaw axis, in addition to the regular wheel \cite{wada_2motor_conv_tire}. Wheels with the ability to drive in both the left and right directions without the need for directional changes have also been developed \cite{tadakuma_dual_rings, honda_uni-cub3}. Both types of wheels can achieve omnidirectional movement, but they require two actuators for each wheel, which presents challenges in terms of weight, volume, price, control, power consumption, wiring, and other aspects, hindering their widespread adoption for various applications.

On the other hand, by adding passive wheels, there have been developments in achieving omnidirectional movement with a single actuator. Examples of such wheels include the Mecanum wheel \cite{patent_mecanum-wheel}, Omni-wheel \cite{Asama_omni-wheel}, and Omni-ball \cite{tadakuma_omni-ball}. These wheels have been successfully used in practical applications such as forklifts \cite{vetex_lift}, Pepper \cite{softbank_pepper}, and RoboMaster S1 \cite{dji_robomaster}. The Omni-wheel, especially the Single Omni-wheel \cite{Asama_single-omni-wheel}, can be made to be thin and lightweight, and to have the same axial contact position. The Mecanum wheel is cost-effective and excels in forward and backward step-climbing capability. To further enhance step-climbing capability, the Spiral Mecanum Wheel \cite{suzumori_spiral_mecanum}, which achieves high step-climbing capability in both the forward and lateral directions, has been proposed. Furthermore, without relying on friction with a wall surface, there have been proposals for structures that deform the wheels \cite{omni-wheg} or structures that pre-divide the wheels into two parts \cite{yoneda-mecanum}, as structures that can be hooked onto steps or other obstacles. This design enhancement has resulted in improved capability for overcoming steps. In this manner, each wheel mechanism possesses its own advantageous characteristics. However, the problem with omnidirectional wheels driven by a single actuator is that the shaft and bearing supporting the free roller become smaller, which leads to decreased strength and increased vulnerability to damage in the event of a collision. Mobilities that move on wheels often experience impact when the wheels collide with surrounding objects during high-speed movement. In cases where they often travel on rough terrain or steps, they frequently experience additional impact associated with falling and landing, in addition to the impact from collisions during driving. One approach to enhancing the strength of omnidirectional wheels is to arrange the spherical contact areas in an alternating pattern, allowing for the use of larger bearings \cite{mourioux_spherical, MY3wheel}. However, a challenge with these arrangements is that they tend to have a greater width despite the improved strength. Among these, the Omni-ball has achieved high strength by reducing the number of bearings and utilizing large bearings, resulting in not only good step-climbing capability but also an overall level of high strength \cite{tadakuma_omni-ball}.

\begin{figure}[tb]
        \centering
        \includegraphics[width = \linewidth]{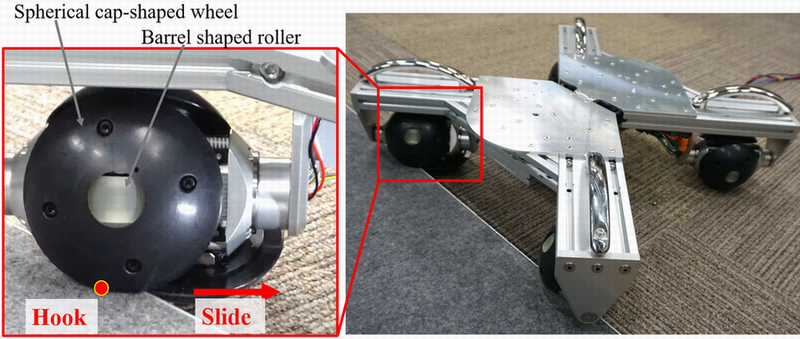}
        \caption{Omnidirectional chassis structure with PTOB can drive even on uneven terrain.}
        \label{fig_main}
\end{figure}

In recent years, the development of leg-wheel robots has been increasing \cite{wang_wheel-legged, bjelonic_ral2019, belli_humanoids2021, hyundai_tiger}. Our team has also been developing legged robots and leg-wheel robots with high step-climbing capability \cite{kamikawa_iros2021, takasugi_arxiv2023}. Based on these experiences, there is a need for wheels that are both stronger and smaller. It is desirable to have in-wheel actuator wheels with sufficient strength to withstand leg movement when mounted on the robot's leg end, without being damaged. The integration of actuators into the wheels is not only beneficial for increasing the design flexibility of mobile platform, but also effective in reducing the overall size, weight, and minimizing the number of components \cite{ryoo_isis2016}.
     
One of the challenges with the existing Omni-ball design is that when placing two large hemispherical structures, the actuator shaft needs to pass between those two hemispheres. This requires careful selection of the actuator to fit inside the sphere, making it difficult to integrate into the wheel. If the actuator is selected to protrude from the sphere, the gap widens and the vibration during movement increases.

Even when the actuator is placed outside without integrating it into the wheel, there is a trade-off between the thickness of the shaft and the gap, making it difficult to increase the strength of the shaft section. To fully utilize the step-climbing capability of the Omni-ball, it requires a significant amount of torque and strength. Furthermore, considering the mobile aspect, the speed of movement is also important, requiring powerful actuators. However, achieving a balanced combination of all these factors in mechanical design can be challenging.

In this study, we propose the Passive Transformable Omni-Ball (PTOB) shown in Fig. \ref{fig_main}, which not only solves these trade-off challenges of the Omni-ball but also further improves off-road performance. To address these challenges, we have introduced the following features as a solution:

- Changing the number of divisions in the Omni-ball from 2 to 3 to achieve in-wheel actuation and minimize the gap for reduced vibration.

- Integrating a sliding structure for the passive wheels to further enhance step-climbing capability.

- Achieving high strength through the use of large-diameter bearings and linear guides at key points.

\section{Design OF Passive Transformable Omni-Ball}

In this chapter, we will present the fundamental configuration of the PTOB, which is an enhanced structure derived from the Omni-ball.

\subsection{In-Wheel Actuator with Three Spherical Cap Structures}

\begin{figure}[tb]
        \centering
        \includegraphics[width = 0.95\linewidth]{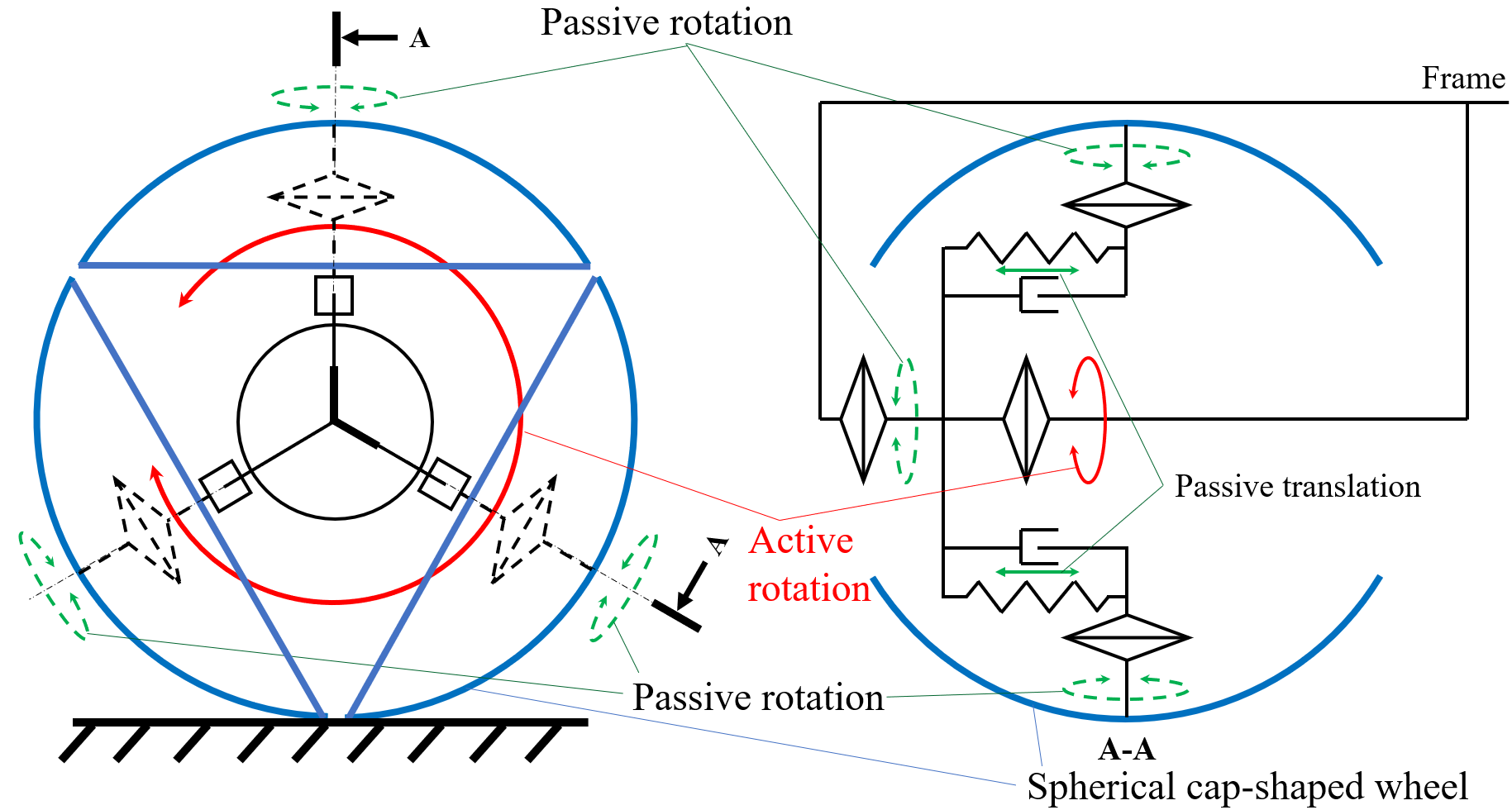}

        (a)
        \vskip\baselineskip
        \includegraphics[width = 0.95\linewidth]{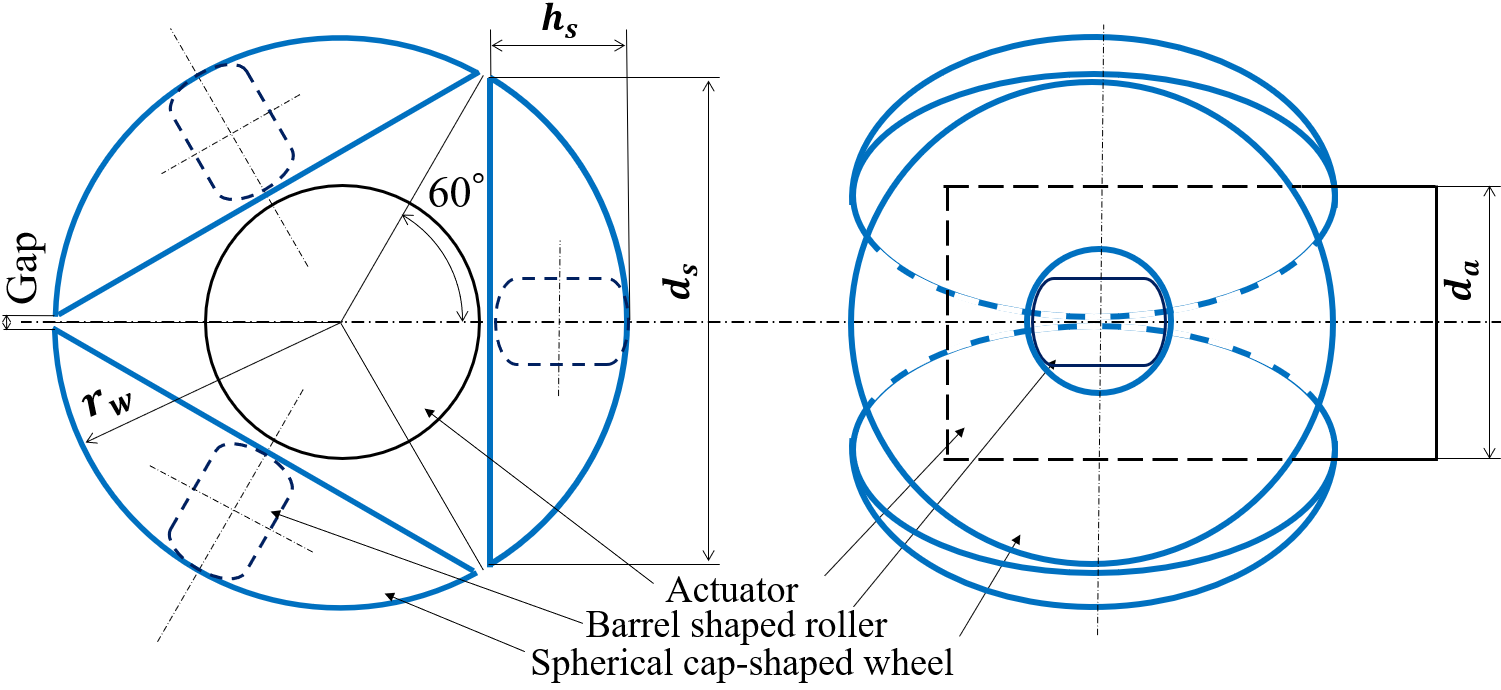}

        (b)
        \vskip\baselineskip
        \includegraphics[width = 0.95\linewidth]{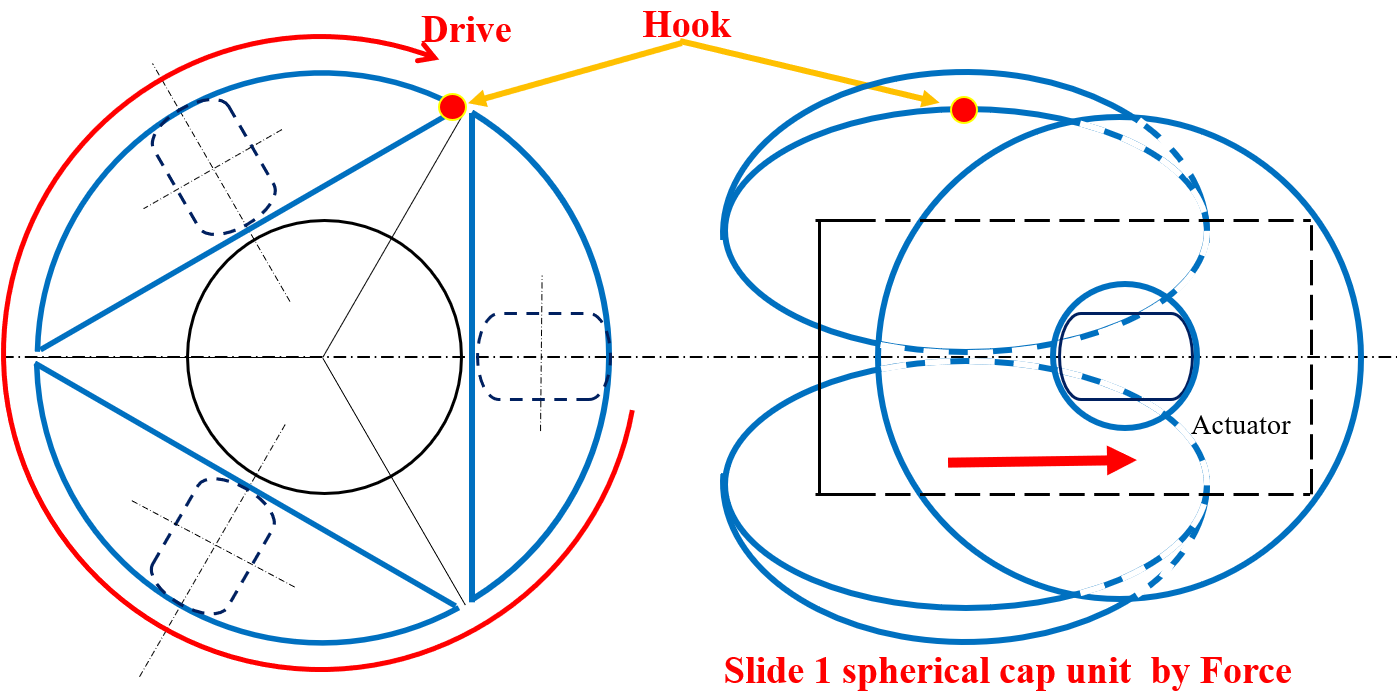}

        (c)
        \caption{Basic structure of the PTOB: (a) Three spherical cap-shaped wheels that contact the ground and have sliding parts that move passively. (b) The spherical cap-shaped wheels have a hollow structure with a barrel-shaped roller inside, covering the singular points of the spherical cap-shaped wheels, similar to the structure of the Omni-ball. (c) The sliding motion allows the wheels to engage with steps by sliding against external forces.}
        \label{fig_ptob_math}
\end{figure}

The basic configuration of the Omni-ball exhibits good characteristics in terms of step-climbing capability and passive wheel strength. While maintaining these features, we have achieved in-wheel integration. To achieve in-wheel integration, it is necessary to arrange a wheel structure with a spherical shape that provides space for inserting the actuator.
The foundational architecture of the PTOB is illustrated in Fig. \ref{fig_ptob_math}. A critical aspect of this design, as Fig. \ref{fig_ptob_math} demonstrates, is the composition of the structure from three spherical cap structures, diverging from the conventional dual-hemisphere configuration. This innovative design yields two significant outcomes:

\begin{itemize}
        \item It facilitates the creation of a space aligned with the rotational axis, which accommodates the integration of the actuator.
        \item It ensures the minimization of the gap between the spherical cap structures to nearly zero, enhancing the wheel's structural coherence.
\end{itemize}

The detailed structure of the spherical cap-shaped wheel and barrel-shaped roller parts themselves are described in the Omni-ball paper \cite{tadakuma_omni-ball}. 

As shown in Fig. \ref{fig_ptob_math}(b), the dimensions of the spherical cap-shaped wheels can be designed and constrained geometrically using parameters such as the wheel radius $r_{w}$. When the gap is zero, the thickness of the spherical cap-shaped wheel is $h_{s} = \frac{1}{2} r_{w}$, and the maximum outer shape of the spherical cap-shaped wheel is $d_s = \sqrt{3} r_{w}$. However, to avoid contact between adjacent spherical cap-shaped wheels, the dimensions need to be smaller.  Therefore, the following constraints apply:

- Thickness of the spherical cap-shaped wheel: 
$h_{s} < \frac{1}{2} r_{w}$.

- Maximum outer diameter of the spherical cap-shaped wheel: 
$d_{s} < \sqrt{3} r_{w}$.

Considering that the actuator is placed to avoid the spherical cap-shaped wheel, assuming a typical cylindrical actuator, the actuator diameter needs to be selected to avoid interference with the thickness of the spherical cap-shaped wheel. This leads to the following constraint:

- Actuator diameter:
$d_{a} < 2(r_{w} - h_{s})$.

\begin{figure}[t]
        \centering
        \includegraphics[width = \linewidth]{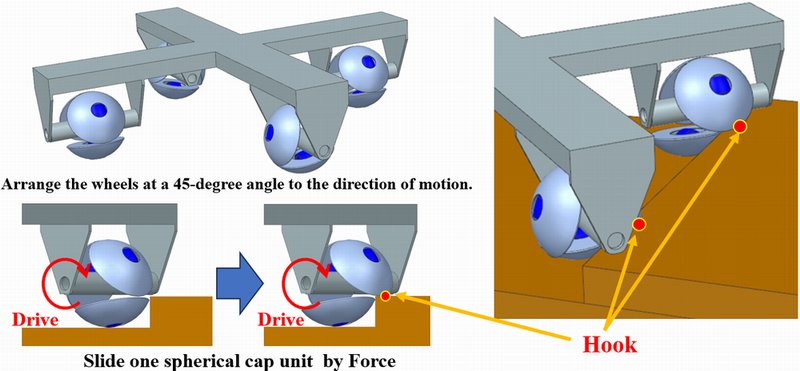}
        (a)

        \vskip\baselineskip

        \includegraphics[width = \linewidth]{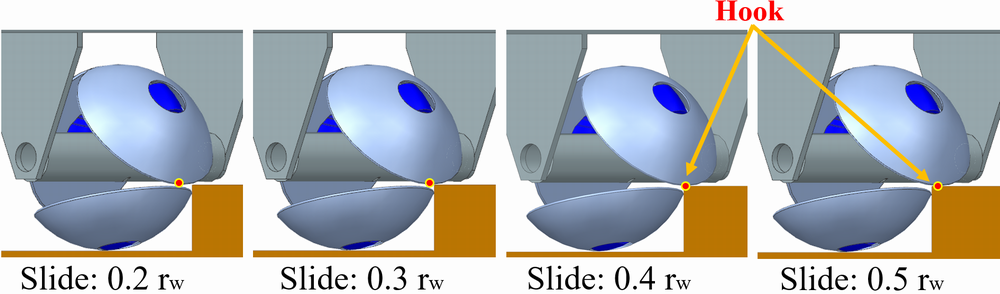}
        (b)
        \caption{Relationship between sliding distance and the height of the step that can be overcome: (a) When the PTOB is positioned at a 45-degree angle to the direction of the step, the spherical cap-shaped wheels come into contact with the step and slide. (b) Whether the step is cleared or not depends on the sliding distance. Approximately 40 $\%$ or more of the radius is required for successful clearance.}
        \label{fig_hook}
\end{figure}

\subsection{Enhanced Step-Climbing Capability with a Sliding Structure}

In order to improve the step-climbing capability of the wheel individually, a sliding structure in the axial direction was introduced to each spherical cap-shaped wheel unit, as shown in Fig. \ref{fig_ptob_math}(a). When the spherical cap-shaped wheel makes contact with a step, external force causes the spherical cap-shaped wheel unit to passively slide, as shown in Fig. \ref{fig_ptob_math}(c), exposing the edge of the next spherical cap structure. This allows for improved step-climbing capability without relying on friction with the step edge, preventing the wheel from getting stuck. After climbing the step, the spherical cap-shaped wheel unit returns to its central position using a compression spring.

Here, we conducted an investigation on the sliding distance required for step-climbing, specifically when the PTOB is positioned at a 45-degree angle to the direction of travel, as shown in Fig. \ref{fig_hook}(a). When the spherical cap-shaped wheel comes into contact with a step riser, it slides in the direction the rotational axis. This sliding motion allows the next spherical cap-shaped wheel to hook its edge onto the step's tread. However, if the spherical surface of the spherical cap-shaped wheel comes into contact with the corner of the step, it slips and falls off without being able to hook its edge onto the step's tread. Therefore, it is important for the edge of spherical cap-shaped wheel to make contact with the step's tread farther away from the corner in order to maintain stability and prevent slipping. The sliding distance required to achieve this was confirmed by gradually changing the sliding distance, as shown in Fig. \ref{fig_hook}(b). It was confirmed that a sliding distance of approximately 40 $\%$ or more of the step height (0.7 $r_{w}$) is necessary.

We designed and prototyped the actual structure to achieve these goals, and Fig. \ref{fig_ptob_design} shows the structure, exploded view, and cross-sectional view. The restorative force of the spherical cap structure is achieved using compression springs.

\subsection{Detailed Design}

In order to improve the step-climbing capability of the wheel individually, we conducted a structural study that included the surrounding outer shell structure of the spherical cap structure.

The Omni-ball and Omni-wheel are often arranged at angles of 90 degrees from each other, with the axial direction positioned at a diagonal angle of 45 degrees to the direction of travel. To ensure that the outer shell structure does not become an obstacle when encountering a step in front, and to facilitate riding over the step when moving diagonally with momentum, the outer shell is designed primarily in a conical shape.

In our design, we aimed to overcome steps that exceed one-third of the wheel diameter, so it was important to avoid situations where the actuator itself gets caught on the step, leading to a decrease in step-climbing capability. Therefore, the diameter of the actuator was determined based on the constraint condition $d_{a} < \frac{2}{3} r_{w}$.

Typically, in a structure like this where the entire periphery of the actuator rotates, an outer rotor motor is a suitable actuator. However, considering the required torque, wheel diameter, and actuator diameter, we chose a geared motor that allows for a smaller actuator diameter. The geared motor is fixed at the back by tightening a screw.

\begin{figure}[!h]
        \centering
        \includegraphics[width = \linewidth]{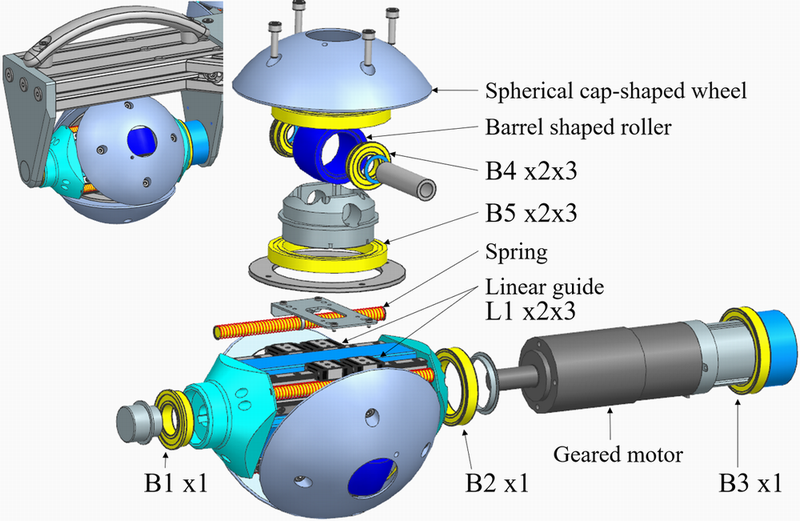}

        \vskip\baselineskip

        \includegraphics[width = \linewidth]{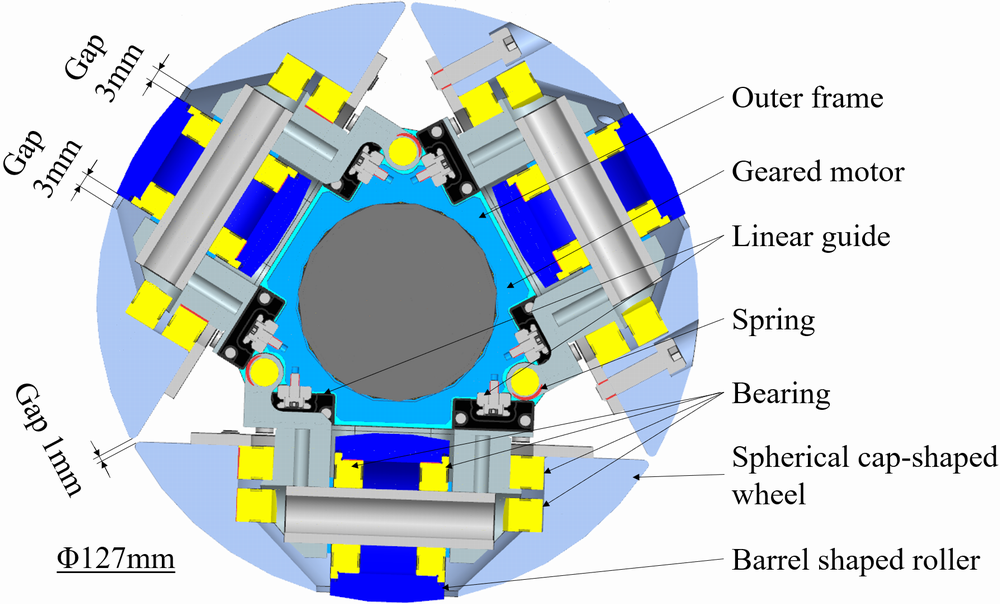}
        (a)

        \vskip\baselineskip

        \scalebox{0.8}[0.8]{
                \begin{tabular}{|c|c|c|}
                \hline
                No. & Type & Specification / product\\
                \hline
                B1 & Bearing & Cr: 3910 N, Cor: 2210 N\\
                \hline
                B2 & Bearing & Cr: 3857 N, Cor: 2892 N\\
                \hline
                B3 & Bearing & Cr: 4020 N, Cor: 3248 N\\
                \hline
                B4 & Bearing & Cr: 2453 N, Cor: 1173 N\\
                \hline
                B5 & Bearing & Cr: 5619 N, Cor: 5177 N\\
                \hline
                L1 &
                \begin{tabular}{c}Liniear Guide\\ (2 blocks) \end{tabular} &
                \begin{tabular}{c}C: 1580 N, Co: 2540 N\\Ma:12.5 Nm, Mb:12.4 Nm,\\Mc:9.2 Nm\end{tabular}\\
                \hline
                \end{tabular}
        }
        \scalebox{0.8}[0.8]{
                \begin{tabular}{cll}
                        *&Cr &: Basic Dynamic Radial Load Rating of Bearing\\
                        *&Cor &: Basic Static Radial Load Rating of Bearing\\
                        *&C &:  Basic Dynamic Load Rating of Linear Guide\\
                        *&Co &:  Basic Static Load Rating of Linear Guide\\
                        *&\multicolumn{2}{l}{Ma, Mb, Mc : Allowable Static Moment}
                \end{tabular}
        }

        (b)
        \caption{(a) External appearance, exploded view, and cross-sectional view of the PTOB. Each spherical cap-shaped wheel is equipped with a linear guide to achieve high strength and sliding.
        (b) Product characteristics of bearings and linear guides. Selected based on prioritizing strength, considering installation at the leg end. The numbers correspond to the exploded view.}
        \label{fig_ptob_design}
\end{figure}

In terms of strength, we aimed for each wheel to be capable of withstanding a maximum instantaneous force of 2,000 N and conducted the selection of sliding components accordingly. The arrangement and performance of the selected sliding components are shown in Fig. \ref{fig_ptob_design}. Both radial and thrust loads need to be supported by the bearings, and they need to rotate smoothly. Therefore, deep groove ball bearings are adopted for both purposes. Bearing No.4 in Fig. \ref{fig_ptob_design}(b) is the weakest, but as depicted in Fig. \ref{fig_ptob_design}, by arranging two identical bearings side by side in a double-end support format, the strength is improved, demonstrating that the wheel as a whole can achieve high strength. Furthermore, we have confirmed through experimental results that the system is able to traverse repeatedly without any issues even when encountering a 45 mm step. The specifications of each component of the PTOB are shown in Table \ref{table_PTOB}.

\begin{table}[!h]
        \caption{Specifications of the PTOB.}
        \label{table_PTOB}
        \begin{center}
        \begin{tabular}{|l|l|}
        \hline
        Wheel diameter & 127 mm\\
        \hline        
        Maximum diameter of & \multirow{2}{*}{36 mm}\\
        barrel shaped roller &\\
        \hline
        \multirow{2}{*}{Material of wheel} & Hard urethane rubber\\
                                                & (Hardness 90)\\
        \hline
        Load capacitiy (Target) & 2000 N\\
        \hline
        Weight / wheel & 2.5 kg\\
        \hline
        \multirow{2}{*}{Actuator} & Maxon : Brushless Motor\\
                                        & EC-i40 70 W 48 V 496656\\
        \hline
        \multirow{2}{*}{Gearbox} & Maxon : Planetary Gearhead\\
                                        & GP42C 26:1 203119\\
        \hline
        Motor driver & Elmo : Gold Whitsle\\
        \hline
        Instant maximum torque & \multirow{2}{*}{11.3 Nm}\\
        (after reduction) &\\
        \hline
        Maximum continuous torque & \multirow{2}{*}{3.9 Nm}\\
        (after reduction) &\\
        \hline
        Maximum speed & \multirow{2}{*}{1260 mm/s (4.5 km/h)}\\
        (after reduction) &\\
        \hline
        Instant maximum & \multirow{2}{*}{178 N / wheel}\\
        propulsive force &\\
        \hline
        Slide spring & SAMINI : 12-0828\\
        \hline
        Range of slide movement & $\pm$ 30 mm\\
        \hline
        Restoring force of slide & Max 12.7 N\\
        \hline        
        \end{tabular}
        \end{center}
\end{table}

\begin{figure}[!h]
        \centering
        \includegraphics[width = \linewidth]{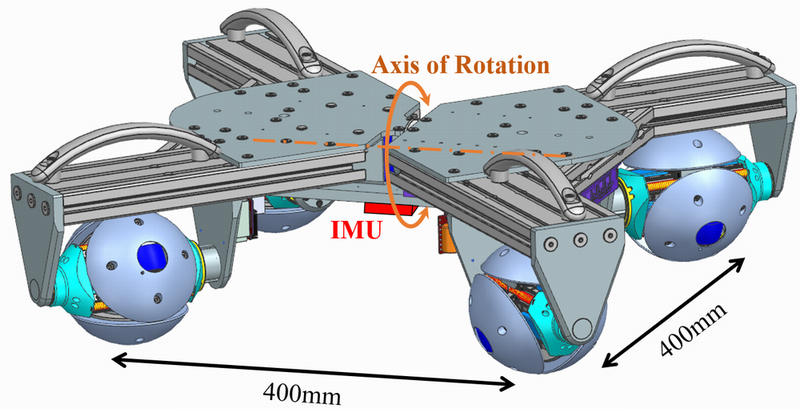}
        \caption{Overall view of the chassis structure. The wheel contact positions are arranged at 400 mm, with a central rotation axis to ensure easy contact with all four wheels even on uneven terrain. An IMU is installed in the center for evaluation purposes.}
        \label{fig_chassis}        
\end{figure}

\begin{table}[!h]
        \caption{Specifications of the chassis structure.}
        \label{table_chassis}
        \begin{center}
        \begin{tabular}{|l|l|}
        \hline
        Number of wheels & 4 (90 deg each)\\
        \hline
        \multirow{2}{*}{Size} & $595 \times 595 \times 207$ mm\\
                                & (Wheel distance : 400 mm)\\
        \hline
        Total weight & 17.6 kg\\
        \hline
        \end{tabular}
        \end{center}
\end{table}

\section{Omnidirectional Wheel Chassis}

We will describe the chassis structure and control system that were prototyped using PTOB. The chassis structure is shown in Fig. \ref{fig_chassis}, and the specifications are presented in Table. \ref{table_chassis}. Four PTOBs are arranged at 90-degree intervals, each with a dual-end support structure as shown in Fig. \ref{fig_support}.

\subsection{Chassis structure}

By incorporating a centrally located passive rotating axis (Fig. \ref{fig_chassis}), the chassis structure allows all wheels to maintain contact with the ground even on uneven terrain.

\begin{figure}[tb]
        \centering
        \includegraphics[width = \linewidth]{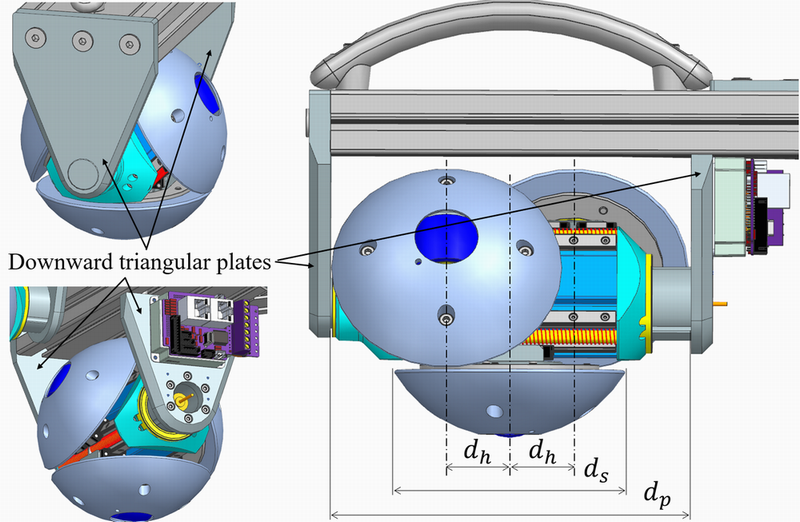}
        \caption{Wheel support spacing of the PTOB. The spherical cap-shaped wheels are positioned with dimensions slightly outside the range of movement of the linear guides to avoid contact with the support plates. The support plates are designed in a downward triangular shape to minimize interference with the environment surrounding the spherical cap-shaped wheels.}
        \label{fig_support}
\end{figure}

The left and right wheel support sections are designed with a downward triangular plate structure (Fig. \ref{fig_support}), reducing the risk of the support sections coming into contact with the environment before the wheels. 

The ideal design for the spacing of downward-facing triangular plates with both ends supported is to minimize the width according to the constraints that do not hinder the range of motion during sliding, as excessive length can lead to decreased strength and increased weight.

(Spacing between support plates)

\begin{equation}
        d_{p} > d_{s} + 2d_{s}
\end{equation}

\subsection{Control System}

\begin{figure}[tb]
        \centering
        \includegraphics[width = 0.8\linewidth]{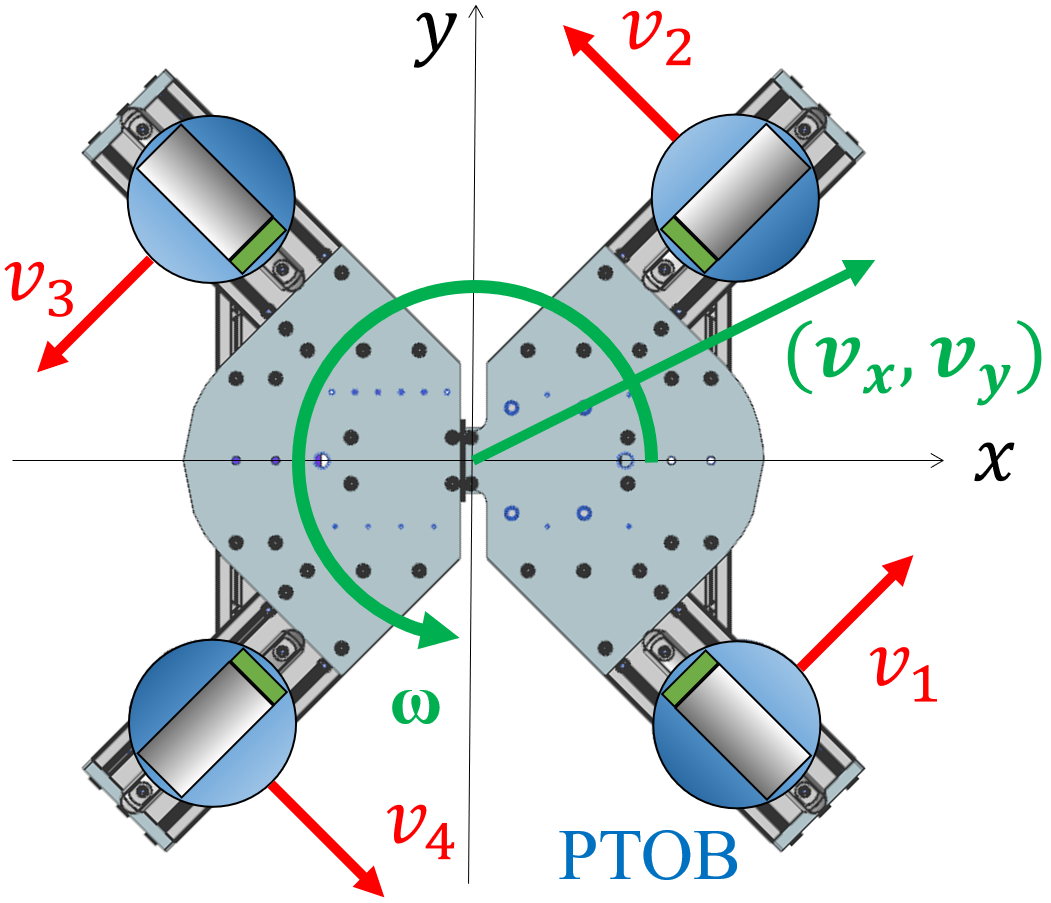}
        \includegraphics[width = 0.95\linewidth]{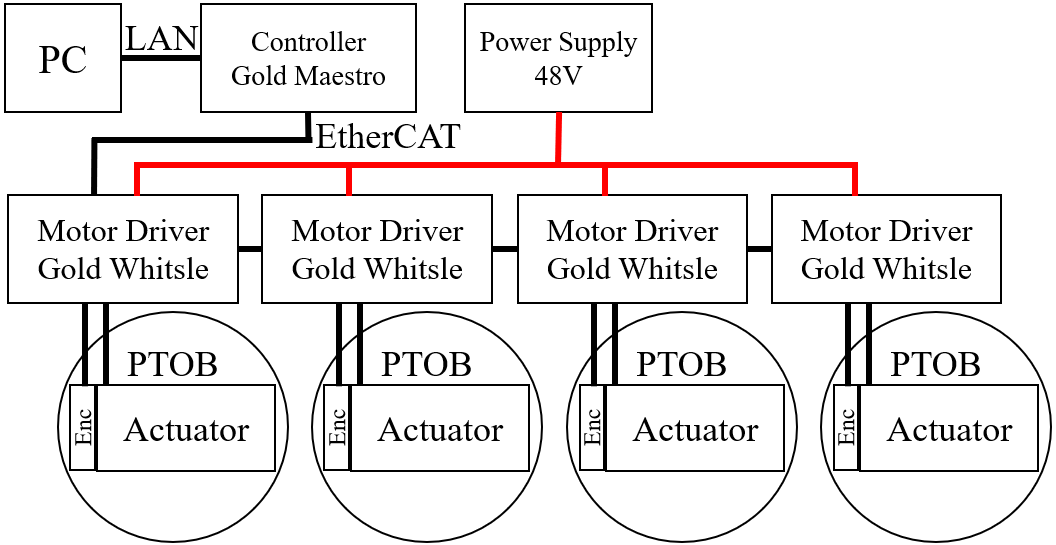}
        \caption{Control system configuration diagram. Each PTOB is equipped with an internal geared motor, which is driven by a motor driver to perform omnidirectional translational and rotational movements.}
        \label{fig_control}
\end{figure}

The drive control method is described here. For control, a PC and Elmo: Gold Maestro are used as shown in Fig. \ref{fig_control}.

The principle of roughly determining the speeds $(v_{1},v_{2},v_{3},v_{4})$ of each wheel to achieve the target velocity $(v_{x},v_{y})$ and rotation rate $(\omega)$ is the same as the conventional Omni-wheel \cite{song_mrmi2006} and Omni-ball \cite{tadakuma_omni-ball}, and is expressed by the following equation.

\begin{equation}
        \left[ \begin{array}{r}
                v_{1} \\ v_{2} \\ v_{3} \\ v_{4}
                \end{array} \right]
                =
                \left[ \begin{array}{rrr}
                1/\sqrt{2} & 1/\sqrt{2} & r \\ -1/\sqrt{2} & 1/\sqrt{2} & r \\ -1/\sqrt{2} & -1/\sqrt{2} & r \\ 1/\sqrt{2} & -1/\sqrt{2} & r
                \end{array} \right]
                \left[ \begin{array}{r}
                        v_{x} \\ v_{y} \\ \omega
                \end{array} \right]
\end{equation}

\section{Experiments}

To confirm the usefulness of the developed omnidirectional wheels, the following experiments were conducted using an omnidirectional platform.

\begin{itemize}
        \item Vibration measurement test during motion on flat ground (compared to the Omni-wheel)
        \item Step-climbing test (compared to the Omni-ball)
        \item Various terrain test
\end{itemize}

\subsection{Vibration measurement test during motion on flat ground}

\begin{figure}[!h]
        \centering
        \includegraphics[width = 0.65\linewidth]{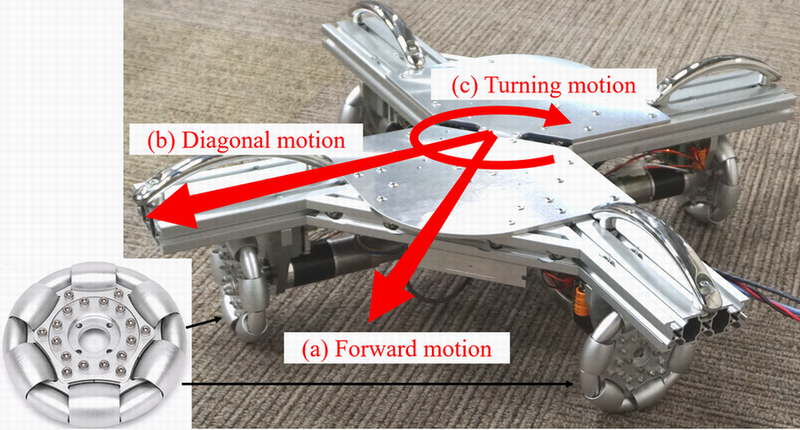}
        \caption{Comparative body with Single Omni-wheel version. The actuators, wheel placement, and wheel diameter were aligned to compare vibrations during three types of movement.}
        \label{fig_dir}

        \vskip\baselineskip

        \centering
        \includegraphics[width = 0.9\linewidth]{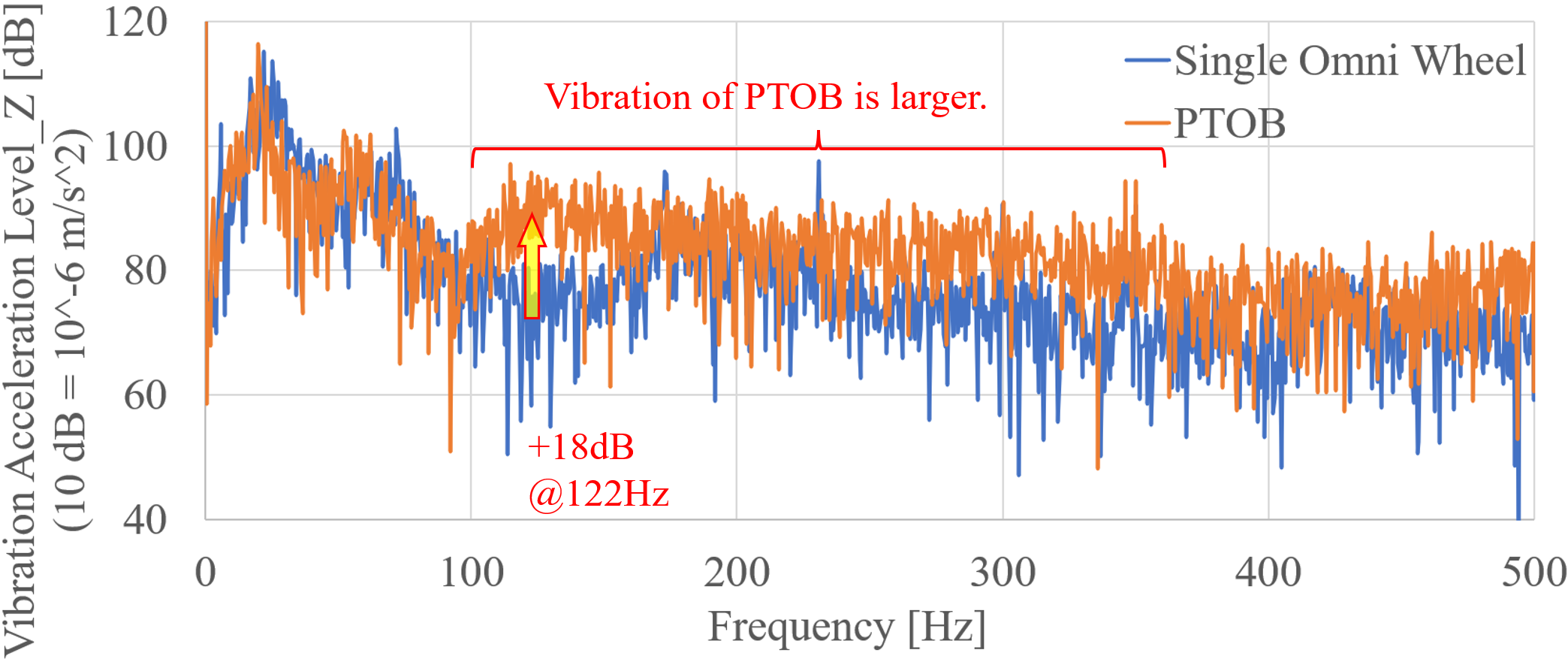}
        (a)Forward motion
        \vskip\baselineskip
        \includegraphics[width = 0.9\linewidth]{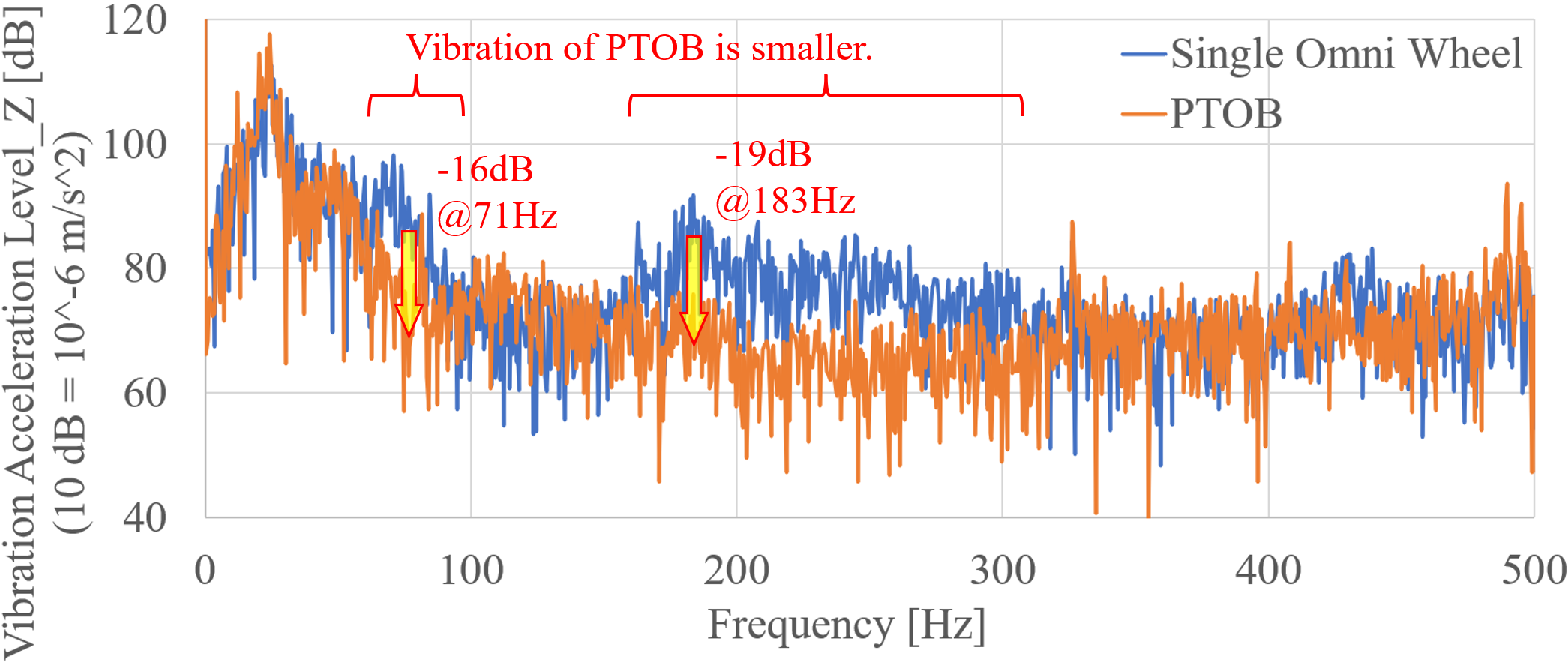}
        (b)Diagonal motion
        \vskip\baselineskip
        \includegraphics[width = 0.9\linewidth]{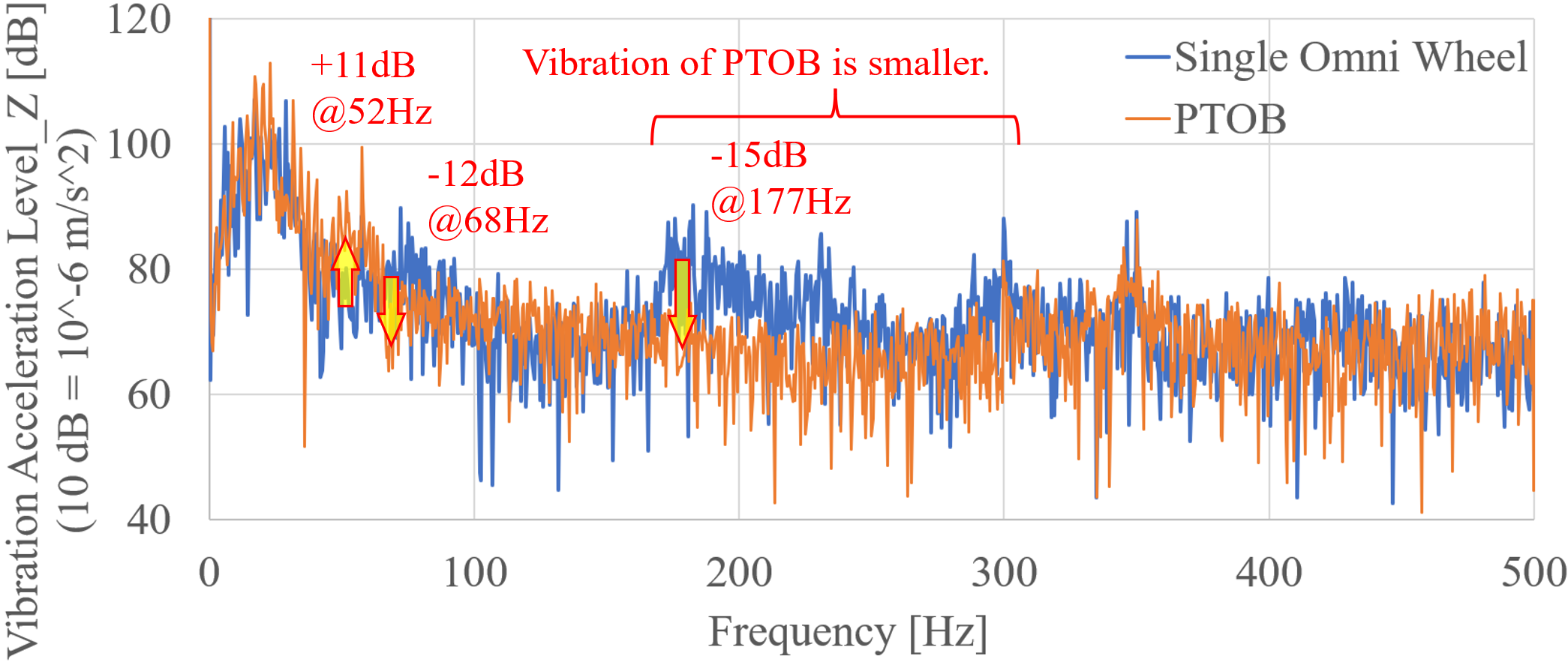}
        (c)Turning motion
        \caption{Vibration acceleration levels after FFT analysis using IMU acceleration data for each movement state. Slight reduction in vibration compared to Omni-wheel observed during diagonal and turning motions. Slight increase in vibration observed during forward and backward movement due to reciprocating sliding motion. Overall, it was confirmed to be at an equivalent level.}
        \label{fig_vib}

        \vskip\baselineskip

        \centering
        \includegraphics[width = \linewidth]{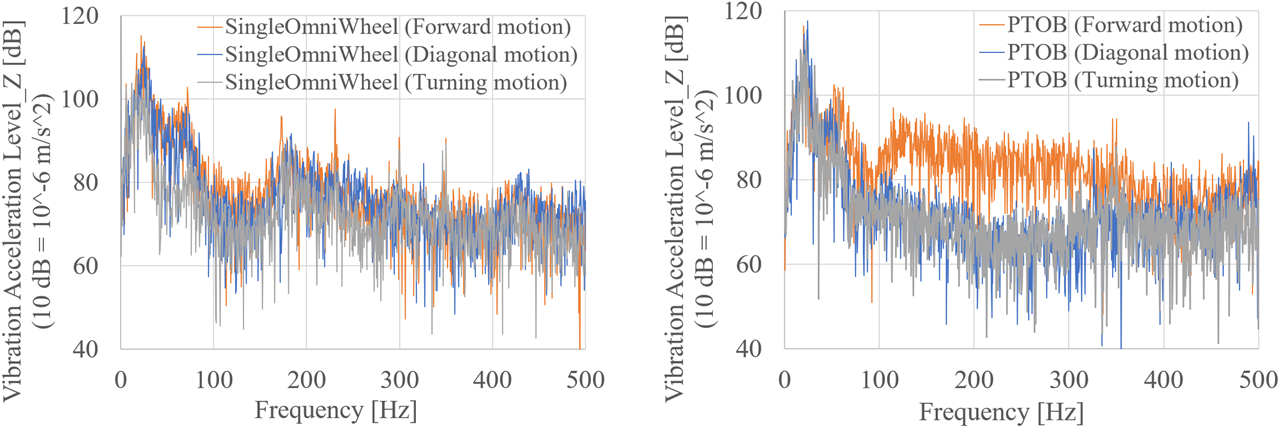}
        \caption{Vibration acceleration levels after FFT analysis using IMU acceleration data of each omnidirectional wheel.}
        \label{fig_each}
\end{figure}

We modified the chassis structure shown in Fig. \ref{fig_chassis} and conducted a comparative experiment by equipping it with commercially available wheels (Vstone: 127 mm Aluminum Single Omni-wheel) that enable low vibration and smooth movement, as shown in Fig. \ref{fig_dir}.

The actuators, gearboxes, encoders, motor drivers, and control system are consistent across models, with an identical wheel placement radius. While the wheel diameters are uniform, the materials vary: Omni-wheels are composed of aluminum, whereas PTOB wheels utilize urethane rubber. It is important to highlight that the Omni-wheels are rated for a maximum load capacity of 200 N per wheel, which is significantly less than the PTOB wheels' intended load capacity of 2000 N per wheel.

The direction of movement in the experiment is shown in Fig. \ref{fig_dir}, and the results are shown in Fig. \ref{fig_vib}.

\begin{figure}[!h]
        \centering
        \includegraphics[width = \linewidth]{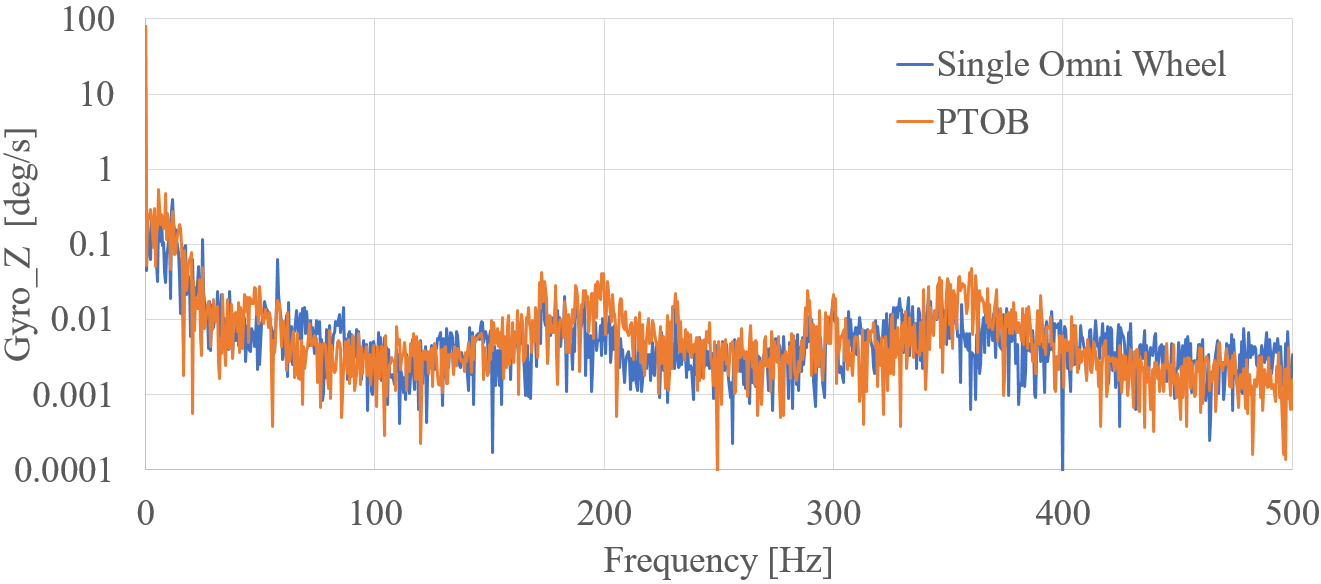}
        \caption{IMU angular velocity data during turning motion. No adverse effects were observed due to the sliding structure.}
        \label{fig_gyro}
\end{figure}

For vibration evaluation, an IMU (Seiko Epson: IMU M-G365) is attached to the bottom center of the vehicle body as shown in Fig. \ref{fig_chassis} and sampled at 1 kHz. In the driving experiments, the rotation speed of the wheels is set to the same value (0.96 rps), and the vibration acceleration levels after FFT analysis are described using the gravity direction acceleration data during each motion. The running location is a tile carpet in the office shown in Fig. \ref{fig_dir} (right), which has minimal irregularities.

Although there are differences in materials, since the rotation speed of the wheels was kept constant, all types of movements showed peaks around 20 Hz in a similar manner.
The variations in vibration due to the difference in wheels are within $\pm$ 19 dB for each motion, indicating that smooth movement was achieved in all cases.

Fig. \ref{fig_each} (left) compares the results for the three motions with the Omni-wheel, while Fig. \ref{fig_each} (right) compares the results for the three motions with PTOB. The Single Omni-wheel shows no dependence on the direction of travel. However, with PTOB, in the case of forward motion, there is an increase in vibration acceleration levels by approximately 20 dB in the frequency range above 50 Hz compared to diagonal and turning motions.

This is due to the sliding structure of the spherical cap-shaped wheel gradually sliding while in contact with the ground, and being returned to the center by the spring's restoring force during the period when the slid spherical cap-shaped wheel is not in contact with the ground. On the other hand, in diagonal and turning motions, the sliding structure does not slide, maintaining its central position, resulting in less vibration.

This phenomenon is also observed during turning motion. It was anticipated that if the sliding structure of the wheels is not centered, the radius at which the driving wheels make contact with the ground may change. However, upon observing the sliding parts during movement, it was observed that they automatically returned to the center due to the force of the spring. This is further confirmed by the angular velocity data during turning motion shown in Fig. \ref{fig_gyro}, which closely matches that of the Single Omni-wheel.

\subsection{Step-climbing test}

The step climbing performance was evaluated by placing steps with a height difference of 5 mm on a flat surface.

First, a step with a height of 25 mm in the diagonal motion in Fig. \ref{fig_dir} was successfully climbed. This is equivalent to the ratio (20 $\%$) of the wheel diameter that was previously successful in climbing with the conventional Omni-ball, confirming that it has equivalent performance.

\begin{figure}[tb]
        \centering
        \includegraphics[width = \linewidth]{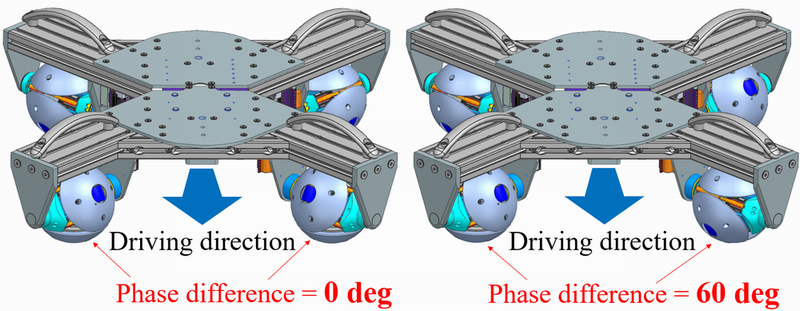}
        \caption{Phase difference between omnidirectional wheels during operation. In omnidirectional wheels based on the Omni-ball, the performance of movement and step climbing ability changes when there is a difference in phase between the left and right wheels.}
        \label{fig_phase}

        \vskip\baselineskip

        \centering
        \includegraphics[width = \linewidth]{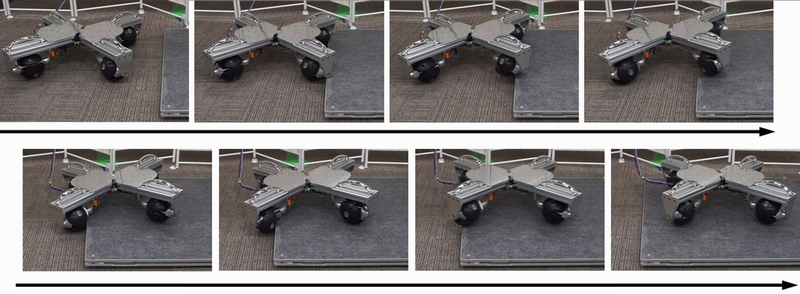}
        \caption{State of overcoming a step (45mm step)}
        \label{fig_step}
\end{figure}

Next, about the forward motion is described. Three stages are compared: the slider part that can freely slide within a range of $\pm$ 30 mm, the one that is restricted within a range of $\pm$ 15 mm, and the one that cannot slide. In the process of this experiment, we confirmed that the phase of the wheel shown in Fig. \ref{fig_phase} is important. Fig. \ref{fig_step} shows the state of overcoming a 45 mm step.

Compared to a case in which the phase of the wheels is aligned as shown in Fig. \ref{fig_phase} (left), it was confirmed that when the phase of the wheels is shifted by 60 degrees as shown in Fig. \ref{fig_phase} (right), the climbing performance decreases and it only sways left and right in the direction of travel when encountering high steps. This is because the spherical cap-shaped wheels generate lateral thrust when they come into contact with a step and rotate. When the phase is aligned, the opposing thrust of both wheels allows the free roller to rotate without releasing force, enabling it to overcome obstacles without swaying left or right.

The above experimental results are summarized in Table \ref{table_step}.
It was demonstrated that restricting the axial slide of the spherical cap-shaped wheel to less than 15 mm results in step climbing performance equivalent to that of the conventional Omni-ball. On the other hand, allowing a wider range of slide, such as 30 mm, has been shown to improve the step climbing performance. Nevertheless, for steps of 50 mm or more, the triangular support plates on both sides of the wheel come into contact with the steps, which prevents the wheels from making contact with the steps. This result confirms that the PTOB is capable of climbing steps up to 35 $\%$ of the wheel diameter. This represents a 20 $\%$ improvement in climbing capability compared to the conventional Omni-ball \cite{tadakuma_omni-ball}, which was only able to climb steps up to 29 $\%$ of the wheel diameter. 

\begin{table}[tb]
        \caption{Successful step climbing height for each driving condition.}
        \label{table_step}
        \begin{center}
        \begin{tabular}{|wl{15mm}|wc{11mm}|wc{11mm}|wc{11mm}|}
        \hline
        \multicolumn{2}{|c|}{Successful step} & \multicolumn{2}{c|}{Phase difference [deg]}\\
        \cline{3-4}
        \multicolumn{2}{|c|}{climbing height [mm]}
            &0 & 60\\
            \hline
            Amount of & 0 &35 &30\\
            \cline{2-4}
            slide possible & $\pm$ 15 & 35 & 30\\
            \cline{2-4}
            [mm]& $\pm$ 30 & 45 & 40\\
        \hline
        \end{tabular}
        \end{center}
\end{table}

\begin{figure}[tb]
        \centering
        \includegraphics[width = \linewidth]{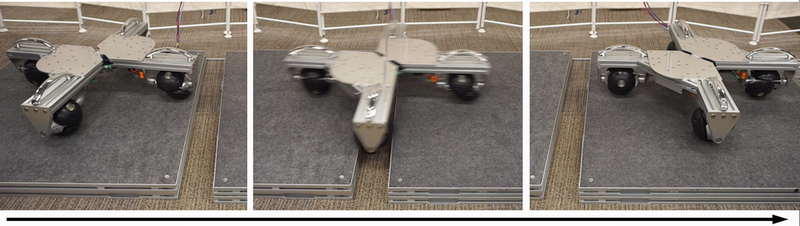}
        \caption{Overcoming a 100 mm gap. Depending on the speed, it is possible to overcome even longer gaps.}
        \label{fig_gap}

        \vskip\baselineskip

        \centering
        \includegraphics[width = \linewidth]{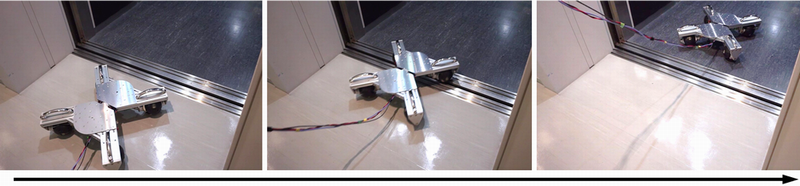}
        \caption{Overcoming the gap in the elevator. We confirmed that stability is maintained during forward, backward, lateral, diagonal, and turning movements.}
        \label{fig_elevator}

        \vskip\baselineskip

        \centering
        \includegraphics[width = \linewidth]{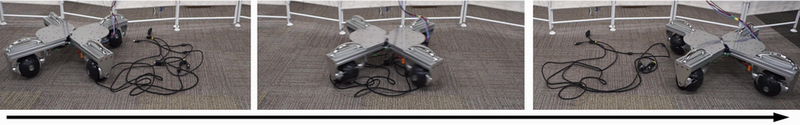}
        \caption{Crossing $\phi$6.5 mm cables. It passed through without any issues, and there was no entanglement of the cables with the wheels.}
        \label{fig_cable}

        \vskip\baselineskip

        \centering
        \includegraphics[width = \linewidth]{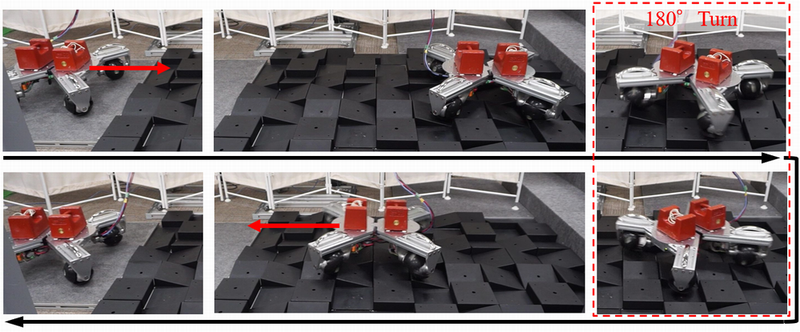}
        \caption{Overcoming uneven terrain, including a 45 mm step. The task involved navigating through uneven terrain with maximum irregularities of 45 mm. The vehicle successfully managed to traverse the terrain by executing entry, 180° rotation, and return maneuvers, requiring careful control.}
        \label{fig_block}
\end{figure}

\subsection{Various terrain test}

To demonstrate the performance of PTOB, driving experiments were conducted in situations with various obstacles.

\begin{itemize}
        \item Gap crossing experiment:\\
        It successfully crossed a gap of 100 mm in both forward and backward directions. This is the same ratio (79 $\%$) relative to the wheel diameter as the success rate achieved with the conventional Omni-ball, confirming the equivalent performance as shown in Fig. \ref{fig_gap}.
        Furthermore, while the success rate decreases, it has also been confirmed that it can cross a gap of 115 mm in both forward, backward, and diagonal directions.
        \item Elevator experiment:
        We confirmed, as shown in Fig. \ref{fig_elevator}, that the developed wheels can smoothly pass through the cargo elevator in either direction (forward motion and diagonal motion) and can also make turns on gaps without any issues.
        \item Cable crossing experiment:
        It successfully crossed a cluster of $\phi$6.5 mm cables placed haphazardly on the ground, as shown in Fig. \ref{fig_cable}, without any issues.
        \item Uneven road traversal experiment:
        It successfully achieved the task of entering, maneuvering, and exiting uneven terrain with steps of 45 mm or less, as shown in Fig. \ref{fig_block}.
        This challenging terrain was composed of a combination of numerous blocks with slope-shaped profiles (each block is 175 mm square).
\end{itemize}

We have successfully achieved crossing in all environments and demonstrated high off-road performance. However, it has been confirmed that there is a possibility of the wheel edges causing damage or entanglement to the wiring during wire crossings. Particular attention is required when dealing with very thin wires and cables.

\section{Conclusion and Future Work}

In this paper, we propose a unique wheel structure based on the Omni-ball, incorporating two modifications: changing the number of divisions to three and adding a passive slider mechanism. We describe the proposed wheel structure and present experimental results, demonstrating its characteristics as a omnidirectional wheel with high step-climbing ability, low vibration, high rigidity, and compact size due to the in-wheel motor.  

In the future, these wheels will be further evaluated for their utility by being attached to wheel-based robots and leg-wheel robots.

On the other hand, we are aware of the following four points as design and operational challenges, indicating that further improvements are necessary for this wheel.

Firstly, In this prototype, significant vibrations occur when climbing steps or moving on uneven terrain. Ideally, not only wheel design but also the use of dampers should be considered to minimize vibrations and prevent them from affecting the transported goods during movement on uneven terrain.

Secondly, In this prototype, we have selected commercially available components such as actuators, encoders, linear guides, bearings, and compression springs without deviating from the range of off-the-shelf products, there may be some strength-related bottlenecks and allowances for the sliding structure to perform large reciprocating movements during straight-line travel. When actually introducing the product, it will be necessary to consider these factors as well.

Thirdly, there is a risk of objects such as cables getting caught in the sliding gaps between multiple spherical cap-shaped wheels, as well as the risk of dirt such as mud adhering to them. Careful operation and the use of covers are necessary to mitigate these risks.

Finally, In order to overcome steps more efficiently, it is necessary to introduce control that takes into account the phase during movement.


\end{document}